%% file: main.tex
\documentclass[10pt,twocolumn]{article}
\usepackage[margin=0.75in,columnsep=0.25in]{geometry}

\usepackage[round]{natbib}

\usepackage{xcolor}
\usepackage{multirow}
\usepackage{hyperref}
\usepackage{url}
\usepackage{booktabs}
\usepackage{graphicx}
\usepackage{amsmath, amssymb}
\usepackage{algorithm}
\usepackage{algpseudocode}
\usepackage{tabularx}
\usepackage{placeins}
\usepackage{authblk}

\begin{document}
\title{\Large\textbf{Data-Efficient Classifier Pretraining Through Diffusion Models}}
\author[1]{Qinyu Li}
\author[1]{Yee Whye Teh}
\author[1]{Chris Holmes}
\author[2, 3]{Joel Jaskari}
\affil[1]{Department of Statistics, University of Oxford}
\affil[2]{Faculty of Medicine and Health Technology, Tampere University}
\affil[3]{Tampere Institute for Advanced Study, Tampere University}
\date{}
\maketitle
\begin{center}\bfseries\Large Abstract\end{center}
Machine learning with limited data is susceptible to overfitting, especially with modern deep neural networks. Prior to the introduction of AlexNet, reconstruction-based pretraining with auto-encoder and denoising auto-encoder were found to be beneficial, partially due to the limited size of the training sets at the time. Motivated by these traditional approaches, we propose DiffPT, a simple diffusion modeling approach to pretrain a given classifier backbone for limited domain-specific training data. We experiment on time-series data, standard natural-image benchmarks, and medical imaging datasets, and contrast the diffusion approach to a multitude of modern and traditional un/self-supervised methods. While contrastive pretraining performs better on CIFAR-10 and Tiny ImageNet in low-data regimes, DiffPT is particularly effective on medical imaging and time-series tasks, where it outperforms self-supervised baselines across a range of low-data regimes. Our results suggest that DiffPT is a simple and effective approach for data-efficient classifier pretraining on different modalities.
\section{Introduction}
\input{01_introduction}
\section{Methodology}
\input{02_methodology}

\section{Related Works}
\input{03_related_works}

\section{Experiments} \label{sec:experiments}
\input{04_experiments}

\section{Results} \label{sec:results}
\input{05_results}
\section{Conclusion}
\input{06_conclusion}
\FloatBarrier
\bibliography{bibliography}
\bibliographystyle{plainnat}
\clearpage
\appendix
\thispagestyle{empty}
\input{07_appendix}

\end{document}

%% file: 01_introduction.tex
There are a variety of domains where deep learning -based classification is of interest, but the development of these models is hindered by the scarcity of data. One such domain is medical and health data, where, e.g., access to data is restricted by legislation, the modality of interest is expensive to acquire, or the medical end-point (disease, complication) is simply rare in the population. In classification, the loss asks the neural network to make only a single categorical decision per data point. However, high dimensional data, e.g., medical imaging and physiological signals, contain substantial amount of information. When labelled data are scarce, many representations will fit the observed labels equally well, and the classification objective alone provides little guidance towards the one that generalises. Training a classifier from scratch can therefore easily overfit the data.

% Unsupervised pretraining in RBMs / deep belief networks (the "good old neural nets")
Unsupervised pretraining used to be common practice for deep neural networks prior to the introduction of AlexNet~\citep{krizhevsky2012imagenet}. Indeed, layer-wise greedy pretraining using deep belief networks \citep{hinton2006fast} or stacked auto-encoders \citep{bengio2006greedy} was found to be crucial in order to obtain high performance with deep neural networks, and was hypothesized to have a regularizing effect~\citep{JMLR:v11:erhan10a}. However, access to larger training sets, along with ReLU activation and better regularization, enabled direct supervised training of these networks~\citep{krizhevsky2012imagenet}.

% Prior non-SSL approaches
Transfer learning based on supervised image classifiers pretrained on the ImageNet~\citep{deng2009imagenet} were found to improve generalization when the fine-tuning domain had only a small amount of data available~\citep{raghu2019transfusion, kornblith2019better, he2019rethinking}. However, the utility of ImageNet transfer learning has been observed to diminish when the fine-tuning modality is dissimilar to natural images, e.g., in the case of chest X-ray imaging~\citep{raghu2019transfusion}. As such, supervised pretraining should be performed on datasets that resemble the modality of the downstream application. Furthermore, it is highly dependent on large labelled datasets, which limits its applicability for modalities with small amounts of data.

%Self-supervised learning (SSL)
Another line of work focuses on self-supervised learning (SSL), where the model is first pretrained on unlabelled datasets, and then fine-tuned on smaller sets of labelled data \citep{chen2020simple, grill2020bootstrap, caron2021emerging}. SSL derives supervisory signals directly from the input data, allowing the representations to be learned without requiring additional manual annotations. However, for effective pretraining, the SSL methods are performed on extremely large unlabelled datasets. In addition, many successful SSL methods rely on carefully designed augmentation strategies. Contrastive methods such as SimCLR~\citep{chen2020simple} construct multiple augmented views of the same image and encourage their representations to agree, while self-distillation methods, such as DINO~\citep{caron2021emerging}, similarly depend on augmentation-induced invariances. These approaches have proved highly successful for natural images, but their effectiveness relies on extensively searching for the optimal augmentations based on the fine-tuning performance. Furthermore, these methods lack a ``standardised blueprint'' on their application for non-imaging modalities, e.g., physiological signals and tabular data.

% Reconstruction based SSL
In contrast, reconstruction-based SSL %objectives 
methods \citep{pascal2008extracting, he2022masked} are often simpler to apply on various modalities, and also resemble traditional unsupervised pretraining -based models. In this work, we revisit the diffusion model \citep{sohl2015deep, ho2020denoising, song2021scorebased} as a pretraining strategy for classifier pretraining, and propose DiffPT (\textbf{Diff}usion \textbf{P}re\textbf{T}raining). Unlike conventional denoising auto-encoders \citep{pascal2008extracting}, which reconstruct images from a fixed level of corruption, diffusion models are trained across a continuum of noise levels, requiring representations that capture both coarse global structure and fine local detail. Because commonly used corruption processes, i.e., the diffusion schedules, have been found to generalize across a variety of domains, DiffPT does not require practitioners to specify domain specific augmentation strategies, unlike in the case of SimCLR and DINO.

Unlike the conventional self-supervised learning setting, which assumes access to a large unlabelled dataset, we consider settings where additional data are scarce or costly to acquire, as is common in medical imaging, scientific experiments, and specialised industrial applications. Motivated by this setting, we ask whether self-supervised pretraining can improve classification even when no additional data are available. We therefore introduce no additional unlabelled data, i.e., the self-supervised pretraining and supervised finetuning are performed on the exactly same dataset in every experiment. We benchmark DiffPT against training from scratch, SimCLR, DINO, auto-encoder, and denoising auto-encoder pretraining with matched architectures. Our contributions are as follows:
\begin{itemize}
\item We propose DiffPT, a data-efficient classifier pretraining method based on diffusion models.
\item We compare DiffPT with a range of self-supervised pretraining methods on various data modalities (physiological signals, natural images, medical imaging) and degrees of data scarcity.
\item We show that DiffPT matches strong augmentation-based self-supervised learning methods on a variety of modalities while requiring less computational resources.
\end{itemize}

%% file: 02_methodology.tex
We consider the problem of learning a classifier model from a limited labelled dataset without access to additional unlabelled data. Let $f_{\theta}$ denote the classifier backbone. Our objective is to improve the representation learned by $f_{\theta}$ through self-supervised pretraining. In the following, we present our diffusion-based pretraining method DiffPT, its implementation details, as well as other standard pretraining strategies that we compare with.

\subsection{DiffPT}

Diffusion models are widely used for generative modelling. Here, we propose to use the diffusion objective to pretrain $f_{\theta}$, which is then finetuned in downstream classification tasks. We therefore treat $f_{\theta}$ as the encoder of the diffusion model. 

For pretraining, we use the denoising diffusion probabilistic model (DDPM) formulation~\citep{ho2020denoising}, which is the discrete-time version of the variance preserving stochastic differential equation~\citep{song2021scorebased}. Let $x_0 \sim p_{\mathrm{data}}$ denote a clean training example. The DDPM forward diffusion process is defined as:
\begin{equation}
    \boldsymbol{x}_t = \sqrt{\bar \alpha_t}\boldsymbol{x}_0 + \sqrt{1 - \bar \alpha_t}\boldsymbol{\epsilon},
\end{equation}
where 
\begin{equation*}
    \alpha_t = 1-\beta_t,\qquad\bar{\alpha}_t = \prod_{s=1}^{t}\alpha_s,
\end{equation*}
and $\{\beta_t\}_{t=1}^{T}$ is the variance schedule. We used the standard linear schedule in \citet{ho2020denoising}.

The diffusion model, $\epsilon_{\theta, \psi}$, is constructed by augmenting the encoder $f_{\theta}$ with a lightweight decoder and skip connections to obtain a U-Net-style architecture. To condition the network on the diffusion timestep, we compute a sinusoidal timestep embedding, transform it with a small multilayer perceptron, and add its linear projections to the output of each encoder block. Here, $\psi$ denotes the parameters of the diffusion-specific components introduced for pretraining. The model is trained with the epsilon prediction objective 
\begin{equation}
    \mathcal{L}_{\mathrm{diff}}
    =
    \mathbb{E}_{\boldsymbol{x}_0,t,\boldsymbol{\epsilon}}
    \left[
        \left\|
            \boldsymbol{\epsilon}
            -
            \epsilon_{\theta,\psi}(\boldsymbol{x}_t,t)
        \right\|_2^2
    \right].
\end{equation}
This objective exposes the backbone to inputs spanning a broad range of corruption levels, from lightly perturbed samples to samples dominated by noise.

As for the decoder, it mirrors the encoder's downsampling stages by progressively upsampling the encoder representation and adding intermediate encoder features through skip connections. For image inputs, we use transposed convolutions for upsampling. For time-series, we interpolate each decoder feature map to the temporal resolution of the corresponding encoder stage before applying a convolutional block. This ensures exact alignment of skip connections for variable and odd sequence lengths. A final convolution projects the decoder output to the input dimension.

In generative tasks, standard data augmentation must be used with care when training diffusion models, since the augmentations can leak into the generated samples~\citep{karras2022elucidating}. However, this concern does not apply in our setting, as we only use the diffusion encoder as a feature extractor. We therefore apply the same augmentations during diffusion pretraining and classifier training. We report the specific augmentations that we used in Section~\ref{sec:experiments}.

\subsection{Baseline methods}
Here we briefly describe the baseline methods used in our experiments. All the methods are adapted to pretrain the same backbone that is pretrained by DiffPT. 

\subsubsection{SimCLR}
SimCLR~\citep{chen2020simple} is a contrastive SSL approach that is based on maximising the similarity of feature representations between augmented views of the same image, while minimizing the similarity between the views of different images. This is achieved through the SimCLR contrastive loss:
\begin{align*}
    &\mathcal{L}_{\text{SimCLR}} = \frac{1}{2N}\sum_{i=1}^N \left( l(\mathbf z^{(1)}_i, \mathbf z^{(2)}_i) + l(\mathbf z^{(2)}_i, \mathbf z^{(1)}_i) \right),\\
    &l(\mathbf z^{(a)}_i, \mathbf z^{(b)}_i) \\
    &= -\log \frac{G(\mathbf z^{(a)}_i, \mathbf z^{(b)}_i)}
    {\sum_{n=1,n\neq i}^NG(\mathbf z^{(a)}_i, \mathbf z^{(a)}_n) + \sum_{m=1}^NG(\mathbf z^{(a)}_i, \mathbf z^{(b)}_m)},\\
    &G(\mathbf z^{(a)}_i, \mathbf z^{(b)}_j) = \exp (\text{sim}(\mathbf z^{(a)}_i, \mathbf z^{(b)}_j))\\
    &\mathbf{z}_i^{(a)} = g(f_{\theta}(t_a(\mathbf x_i))),
\end{align*}
where $\boldsymbol{x}_i$ is a training example from the set of $N$ images, $t_1, t_2$ are two (independent and stochastic) augmentation operators, $f_{\theta}$ is the encoder being pretrained, $g$ is a projection head and sim denotes a similarity measure. We used cosine similarity similar to the original work~\citep{chen2020simple}.

In~\citet{chen2020simple}, it was shown that SimCLR is sensitive to the choice of the set of augmentations induced by $t_1$ and $t_2$. This means that in order for the practitioner to implement the algorithm for their domain, an extensive hyperparameter search must be conducted to find the best augmentations and their transformation magnitudes.

\subsubsection{DINO}
DINO~\citep{caron2021emerging} is an SSL approach based on self-distillation. A \textit{student} model is trained to match the representations of a \textit{teacher} model, where the teacher model's parameters are updated through exponential moving average (EMA) of the student's parameters. The loss function of DINO is defined as:
\begin{align*}
    \mathcal{L}_{\text{DINO}} &= -\frac{1}{NM}\sum_{i=1}^N\sum_{\boldsymbol x_t \in T_i}\sum_{\boldsymbol x_s \in S_i \backslash \boldsymbol x_t}p_t(\boldsymbol{x}_t)\log p_s(\boldsymbol{x_s}),\\
    p_s(\boldsymbol x) &= \text{Softmax}\left(\frac{g_s(\boldsymbol x)}{\tau_s}\right),\\
    p_t(\boldsymbol x) &= \text{Softmax}\left(\frac{g_t(\boldsymbol x) - \boldsymbol m}{\tau_t}\right),
\end{align*}
where $g_s$ and $g_t$ are the student and teacher networks, respectively, with the same architecture consisting of a backbone and a projection head, and $\tau_s$ and $\tau_t$ are the softmax temperatures used by the student and teacher, respectively. $T_i$ is a set of ``global" augmented views of an image $\boldsymbol x_i$, while $S_i$ contains both the $T_i$ and multiple ``local" augmented views, i.e., crops, of $\boldsymbol x_i$. The constant $M$ is the number of evaluations in the two innermost sums for computing the average, and the vector $\boldsymbol m$ is a centering parameter used for the teacher logits, estimated with EMA while training. The logit temperature and centering are used to avoid model collapse.

We note that the augmentation settings in DINO have an effect on the representation quality, similar to SimCLR. Indeed, in the seminal work, it was found that the optimal lower bound on the sampled global and upper bound on the sampled local cropping range was 0.3~\citep{caron2021emerging}. 

\subsubsection{Auto-encoder approaches}
Traditional auto-encoders (AE) are encoder-decoder networks that typically utilise a feature bottleneck between them to avoid trivial encoding, e.g., identity transform, while the denoising auto-encoder additionally corrupts the input~\citep{JMLR:v11:erhan10a}. The denoising auto-encoder (DAE) resembles a diffusion model in that it is trained to recover the clean input from a noised one~\citep{pascal2008extracting}, however, the noise level is fixed. Indeed, the denoising auto-encoder objective is equivalent to score matching over a specific noise level~\citep{pascal2011connection}. Similar to diffusion models, denoising auto-encoders do not require a bottleneck layer, as the denoising objective requires the model to remove noise even when overcomplete layers are used~\citep{JMLR:v11:erhan10a}.

Similar to DiffPT, the auto-encoder uses the classifier backbone as the encoder network, followed by a simple decoder, but without skip connections between the encoder and the decoder to introduce an information bottleneck. The mean squared error (MSE) loss is used as the loss function:
\begin{equation*}
    \mathcal{L}_{AE} = \mathbb{E}_{\boldsymbol x}\left[||\boldsymbol{x} - d_{\phi}(f_\theta(\boldsymbol{x}))||^2  \right],
\end{equation*}
where $d_{\phi}$ is the lightweight decoder. Denoising auto-encoder is implemented similarly, but the preprocessed inputs have isotropic Gaussian noise added to them, while the network is trained to reconstruct the clean input. Specifically the denoising auto-encoder loss is defined as:
\begin{equation*}
    \mathcal{L}_{DAE} = \mathbb{E}_{\boldsymbol x,\boldsymbol \epsilon}\left[||\boldsymbol{x} - d_{\phi}(f_\theta(\boldsymbol{x} + \boldsymbol{\epsilon}))||^2  \right],
\end{equation*}
where in our experiments $\boldsymbol \epsilon\sim\mathcal{N}(\boldsymbol 0,\sigma^2 I)$ with $\sigma=0.1$.

%% file: 03_related_works.tex
Recent works have explored diffusion models for representation learning by taking an off-the-shelf diffusion model or modifying a part of it. For instance, \citet{li2023your} uses a conditional diffusion model for classification by treating it as a conditional density estimator. \citet{xiang2023denoising} interprets the diffusion models as level-conditional denoising auto-encoders. Similarly, \citet{yang2023diffusion} studies the representations learned by the diffusion model at different timesteps and proposes to select the optimal timestep for knowledge transfer through reinforcement learning. \citet{hybvit} connects diffusion modeling with discriminative tasks using a different approach, training a hybrid model, with the vision transformer architecture, using the diffusion objective jointly with the classification objective.

A range of studies also explore auxiliary encoder approaches by employing an auxiliary encoder to extract representations from clean inputs to aid in the diffusion process. In particular, \citet{drl2} and \citet{drl} extract the infinite-dimensonal representation induced by diffusion. \citet{diti} learns $t$-specific features to make up for the reconstruction errors incurred during diffusion training. \citet{dbae} learns a latent representation of the input by building an information bottleneck. 

Our method is most closely related to \citet{xiang2023denoising}, as both perform diffusion pretraining followed by model truncation and discriminative fine-tuning. However, while \citet{xiang2023denoising} derives the downstream model from standard DDPM architectures, we use the target discriminative backbone as the diffusion encoder, enabling diffusion pretraining of a task-specific architecture. Moreover, while diffusion models for generation typically rely on large-scale training data, our focus on downstream discriminative performance allows us to study diffusion pretraining in small-data regimes, which, to the best of our knowledge, have not been systematically explored.

%% file: 04_experiments.tex
In this section, we will first present our experimental setups, including the datasets, evaluation measures, and pretraining and fine-tuning protocols.

\subsection{Datasets}
\begin{figure*}[h!]
    \centering
    \includegraphics[width=\linewidth]{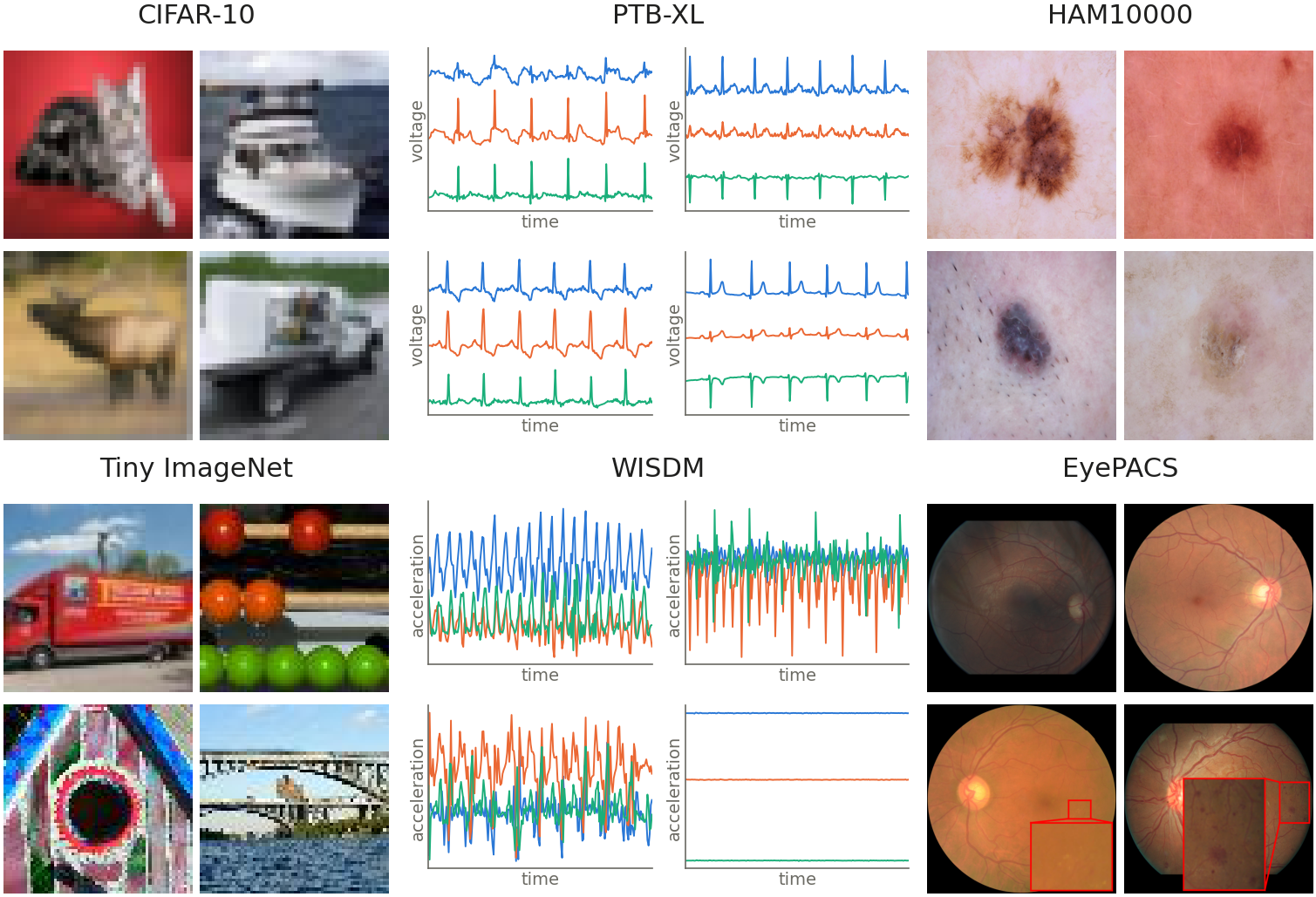}
    \caption{Examples from the datasets used in our experiments. CIFAR-10 and Tiny ImageNet show images from four randomly chosen classes. PTB-XL shows examples from four diagnostic superclasses (truncated to only ECG leads I–III for visualisation purposes). WISDM shows tri-axial phone accelerometer signals from four activities. For EyePACS the top row images are randomly selected, while the bottom row images are selected to present biomarkers suitable for visualisation. The red boxes in the bottom row are zoomed-in regions containing discriminative biomarkers, highlighting their extreme localised nature.}
    \label{fig:datasets}
\end{figure*}

We consider 6 datasets consisting of natural images, time-series, and medical imaging. A visualization of representative examples from each dataset is presented in Figure~\ref{fig:datasets}.

\paragraph{Natural images}
We use two natural image benchmark datasets: CIFAR-10~\citep{krizhevsky2009learning} and Tiny ImageNet~\citep{le2015tiny}. CIFAR-10 contains images from 10 classes with a spatial shape of 32~$\times$~32. Tiny ImageNet is a subset of the ImageNet~\citep{deng2009imagenet} containing 200 classes of 64~$\times$~64 images. For Tiny ImageNet, we use the official 10k-image validation set as our test set. For both datasets we hold out a stratified 20\% of the training set for validation. 

\paragraph{Time-series}
PTB-XL~\citep{wagner2020ptb} contains 21,799 10-second 12-lead clinical ECGs. We use the 100 Hz version and the 5 diagnostic superclasses as a multi-label task, and drop records with no diagnostic superclass label. We use the recommended patient-wise stratified folds: folds 1–8 for training, fold 9 for validation and fold 10 for testing. Following~\citet{strodthoff2021deep}, models are trained on random 2.5~s crops (250 samples). WISDM~\citep{wisdm_smartphone_and_smartwatch_activity_and_biometrics_dataset__507} contains accelerometer and gyroscope recordings from a smartphone and a smartwatch worn by 51 subjects performing 18 activities for about 3 minutes each. We resample all four tri-axial streams onto a common 20~Hz grid, which gives 12 channels. We then split by subject for train/val/test (616/144/141 recordings). Training uses random 10~s (200-sample) windows.

\paragraph{Medical imaging}
We use two medical imaging benchmark datasets: the Kaggle EyePACS~\citep{cuadros2009eyepacs} and HAM10000~\citep{tschandl2018ham10000}. The EyePACS dataset contains retinal fundus photographs of diabetic retinopathy patients. It has 35,126 training images, 10,906 validation images, and 42,670 test images. The original annotation scheme follows 5-class classification, which we cast as binary classification (moderate or worse diabetic retinopathy), following previous works~\citep{gulshan2016development, leibig2017leveraging}. We preprocess the images following~\citet{jaskari2022uncertainty}. HAM10000~\citep{tschandl2018ham10000} contains 10,015 dermoscopic images of pigmented skin lesions in 7 diagnostic categories. Several images can show the same physical lesion, so we split the 10,015 images 80/20 into train and validation grouped by lesion ID to avoid leakage. We test on the ISIC~\citep{codella2019skin} data with 1,511 images. For both datasets, images are resized to $256 \times 256$.

\subsection{Evaluation protocols}
To simulate low data regimes, we train using a subsampled fraction $p$ of each dataset's training split. {Importantly, the same subset is used for self-supervised pretraining and for supervised fine-tuning. Pretraining therefore does not have access to additional unlabelled data. The exact values of $p$ differ for different datasets by sample size: for CIFAR-10, Tiny ImageNet, and EyePACS, we use 1\% and 10\%, for HAM, we use 10\%, for WISDM 50\%, and for PTB-XL, 5\% and 10\%. In addition, we present the full data (100\%) results for every dataset and method.

Since CIFAR-10 and Tiny ImageNet have no official validation split, a class-stratified subset is drawn first, and then 20\% of it is held out as validation, also in stratified manner. PTB-XL and WISDM are subsampled in a similar way, but uniformly at random. EyePACS and HAM10000 have an official separate validation split. That split is subsampled independently by the same fraction $p$, also stratified. We chose to use reduced validation data in reduced data settings, so that model selection does not use more labelled data than training, which is typically the case in real low data scenarios. Test sets are always used in full for precise evaluation of the generalization. We repeat each experiment over 3 seeds, each drawing a different subset, and report the mean. This captures variability from both the choice of subset and optimisation.

With DiffPT, auto-encoder, and denoising auto-encoder, the best pretrained model is selected based on the validation loss, such that the checkpoint selection is label-free for the generative models. For SimCLR and DINO, the training was less stable and validation loss was not a reliable guide to representation quality, as contrastive losses can be only loosely related to downstream performance~\citep{saunshi2022understanding}, and self-distillation losses are computed against a moving teacher, making them hard to compare across checkpoints. We therefore select these models by k-NN accuracy (macro-averaged AUROC for PTB-XL) on frozen validation features.

As for the classifier fine-tuning that follows the pretraining, the best checkpoint is selected by the validation accuracy, or validation macro-averaged AUROC in the case of PTB-XL. For the time-series datasets, we predict on 50\%-overlapping windows and aggregate them by taking the maximum in the case of PTB-XL, following~\citet{strodthoff2021deep}, while in the case of WISDM, we average the prediction. As the evaluation measure, we use accuracy for CIFAR-10, Tiny ImageNet, and WISDM, AUROC for EyePACS, and macro-averaged AUROC for HAM and PTB-XL.

\subsection{Implementation details}
For CIFAR-10 and Tiny ImageNet we use ResNet-18~\citep{he2016deep}, and for EyePACS and HAM10000 we use ResNet-34. We use the CIFAR-style stem in these models, i.e., kernel size of $3 \times 3$, stride 1, and no max-pool, which prevents reducing spatial resolutions too rapidly that can degrade performance on the medical datasets with small lesions. As for the time-series datasets (PTB-XL and WISDM) we use a one-dimensional adaptation of the ResNet, following the implementation in \citet{wang2017time}.

DiffPT, SimCLR, auto-encoder and denoising auto-encoder are trained with AdamW, with a linear learning-rate warm-up followed by cosine decay, and gradient-norm clipping at 1.0. The classifier training in all settings follow a similar recipe. The implementation of DINO follows \citet{caron2021emerging}. Details of data augmentation are provided in Appendix~\ref{app:augmentations}. However, it should be noted that all methods share the same augmentations, specific to each dataset, with the exception of SimCLR and DINO, which have pretraining specific augmentations. As the performance of SimCLR and DINO depends on constructing augmented views that preserve task-relevant discriminative signals, we conducted an augmentation hyperparameter search for these methods on the medical imaging datasets, where discriminative features can be highly localized and thus particularly sensitive to augmentation choices.

All models are trained with (binary/multi-class) cross-entropy loss, while also employing weighting for the imbalanced datasets (HAM10000, EyePACS, PTB-XL) Specifically, let $N$ be the number of training examples, $C$ the number of classes, and $n_c$ the number of training examples in class $c$. We weight the cross-entropy term of each example in class $c$ by
\[
  w_c = \frac{N}{C\,n_c},
\]
so that every class contributes equally to the loss and the average per-example weight is one. PTB-XL is a multi-label task over the five diagnostic superclasses. We use the binary cross-entropy for each superclass $c$, with the positive term weighted by $n^-_c / n^+_c$, where $n^+_c$ and $n^-_c$ are the numbers of training records with and without the label $c$. All weights are computed from the training set of each run. Specifically, in the reduced-data experiments they are computed on the random training set subsample. The experiments were conducted using a range of GPU architectures, including NVIDIA A40 and GH200, and AMD MI250X.

%% file: 05_results.tex
In this section, we will first present the classification results of all the pretraining approaches benchmarked against training from scratch. We will then turn to the computational complexity comparison of the methods.

\subsection{Self-supervised pretraining results}

\paragraph{Natural image classification} On the two benchmarks of small natural images, SimCLR turns out to have the best performance in the subsampled data regimes with DiffPT ranking the second. Interestingly, in the full data regime, DINO turns out to have the best performance out of the pretraining methods for both datasets, being on par with DiffPT for Tiny ImageNet. DiffPT ranks the second on full CIFAR-10 dataset. However, on the full data regime, training from scratch turns out to be on par with DINO pretraining for CIFAR-10, while outperforming pretraining on Tiny ImageNet. Full results are shown in Table~\ref{tab:cifar_tin_acc}.
\begin{table}[!h]
\centering
\caption{Accuracy (\%) on CIFAR-10 and Tiny ImageNet across subset sizes. \textbf{Bold} indicates best, \underline{underline} indicates second-best per column among pretraining methods. AE: auto-encoder, DAE: denoising auto-encoder.}
\label{tab:cifar_tin_acc}
\resizebox{\linewidth}{!}{%
\begin{tabular}{lccc ccc}
\toprule
 & \multicolumn{3}{c}{CIFAR-10} & \multicolumn{3}{c}{Tiny ImageNet} \\
\cmidrule(lr){2-4} \cmidrule(lr){5-7}
Method & 1\% & 10\% & 100\% & 1\% & 10\% & 100\% \\
\midrule
AE  & $44.2$ & $75.5$ & $92.4$ & $5.3$ & $27.7$ & \underline{$57.8$} \\
DAE & $44.4$ & $75.9$ & $92.3$ & $5.8$ & $27.5$ & $57.5$ \\
DINO          & $42.9$ & $76.0$ & $\boldsymbol{93.1}$ & $5.2$ & $27.0$ & $\boldsymbol{58.2}$ \\
SimCLR        & $\boldsymbol{54.3}$ & $\boldsymbol{79.1}$ & $91.2$ & $\boldsymbol{8.8}$ & $\boldsymbol{30.8}$ & $54.9$ \\
DiffPT (ours)    & \underline{$49.7$} & \underline{$78.7$} & \underline{$93.0$} & \underline{$7.6$} & \underline{$29.5$} & $\boldsymbol{58.2}$ \\
\midrule
From scratch        & $42.4$ & $76.2$ & $93.1$ & $4.8$ & $27.0$ & $58.4$ \\
\bottomrule
\end{tabular}%
}
\end{table}

\paragraph{Time-series classification} On WISDM, DiffPT yields substantial accuracy improvements over other pretraining methods and training from scratch when using 50\% and 100\% of the training data. On PTB-XL, DiffPT achieves the best performance among the pretraining methods in the low-data regimes (5\% and 10\%), albeit with a small margin. When using the full PTB-XL training set, however, all pretraining methods perform similarly to training from scratch, suggesting that the benefits of pretraining diminish as more data become available for this dataset. Interestingly, on both datasets, reconstruction -based pretraining methods outperform SimCLR and DINO, with the exception of DINO obtaining the best performance for PTB-XL for 100\% data. Full results are shown in Table~\ref{tab:wisdm_ptbxl_results}.
\begin{table}[!h]
\centering
\caption{WISDM accuracy (\%) and PTB-XL macro-averaged AUROC across subset sizes. \textbf{Bold} indicates best, \underline{underline} indicates second-best per column among pretraining methods. AE: auto-encoder, DAE: denoising auto-encoder.}
\label{tab:wisdm_ptbxl_results}
\resizebox{\linewidth}{!}{%
\begin{tabular}{lcc ccc}
\toprule
 & \multicolumn{2}{c}{WISDM (Acc \%)} & \multicolumn{3}{c}{PTB-XL (AUROC)} \\
\cmidrule(lr){2-3} \cmidrule(lr){4-6}
Method & 50\% & 100\% & 5\% & 10\% & 100\% \\
\midrule
AE  & \underline{$58.16$} & $70.69$ & \underline{$0.869$} & \underline{$0.880$} & \underline{$0.927$} \\
DAE & $57.45$ & \underline{$72.34$} & \underline{$0.869$} & \underline{$0.880$} & \underline{$0.927$} \\
DINO          & $56.50$ & $70.69$ & $0.859$ & $0.870$ & $\boldsymbol{0.928}$ \\
SimCLR        & $55.08$ & $69.98$ & $0.862$ & $0.871$ & $0.926$ \\
DiffPT (ours)    & $\boldsymbol{59.81}$ & $\boldsymbol{74.00}$ & $\boldsymbol{0.872}$ & $\boldsymbol{0.881}$ & \underline{$0.927$} \\
\midrule
From scratch        & $56.03$ & $70.92$ & $0.864$ & $0.874$ & $0.927$ \\
\bottomrule
\end{tabular}%
}
\end{table}

\paragraph{Medical image classification} As for the results for HAM10000, SimCLR has the best performance, while DiffPT ranks the second. All pretraining methods outperform training from scratch. On EyePACS, DiffPT performs the best across all three data regimes (1\%, 10\%, and 100\%), while interestingly denoising auto-encoder has the second best performance for subsets with 1\% and 10\% of data. Notably, with the exception of DINO, all pretraining methods substantially outperform training from scratch in terms of AUROC, even when the full training set is available. Full results are shown in Table~\ref{tab:HAM10000_retino_auroc}.
\begin{table}[!h]
\centering
\caption{HAM10000 (macro-averaged) and EyePACS AUROC across subset sizes. \textbf{Bold} indicates best, \underline{underline} indicates second-best per column among pretraining methods. AE: auto-encoder, DAE: denoising auto-encoder.}
\label{tab:HAM10000_retino_auroc}
\resizebox{\linewidth}{!}{%
\begin{tabular}{lcc ccc}
\toprule
 & \multicolumn{2}{c}{HAM10000} & \multicolumn{3}{c}{EyePACS} \\
\cmidrule(lr){2-3} \cmidrule(lr){4-6}
Method & 10\% & 100\% & 1\% & 10\% & 100\% \\
\midrule
AE  & $0.801$ & $0.891$ & $0.623$ & $0.746$ & $0.890$ \\
DAE & $0.804$ & $0.889$ & \underline{$0.660$} & \underline{$0.840$} & $0.915$ \\
DINO          & $0.797$ & $0.893$ & $0.654$ & $0.659$ & $0.827$ \\
SimCLR        & $\boldsymbol{0.855}$ & $\boldsymbol{0.925}$ & $0.643$ & $0.825$ & \underline{$0.941$} \\
DiffPT (ours)    & \underline{$0.828$} & \underline{$0.905$} & $\boldsymbol{0.744}$ & $\boldsymbol{0.854}$ & $\boldsymbol{0.946}$ \\
\midrule
From scratch        & $0.782$ & $0.887$ & $0.618$ & $0.691$ & $0.829$ \\
\bottomrule
\end{tabular}%
}
\end{table}

\begin{table*}[h!]
\caption{Computational cost of a training step for each pretraining method on the HAM10000 dataset. The batch size is 128 and inputs are $256{\times}256$ RGB images. Using 32-bit floating-point precision and a single NVIDIA GH200 GPU. All methods share the same ResNet-34 backbone (21.3M parameters), which is the only part transferred to the downstream classifier. \emph{Extra params} are trainable parameters used only during pretraining (decoder, projector, or DINO heads). Memory is peak allocated GPU memory; throughput counts source images per second (excluding CPU data loading but including method specific augmentation). Relative values are with respect to the classifier baseline trained from scratch. AE: auto-encoder, DAE: denoising auto-encoder.}
\label{tab:compute}
\centering
\vspace{1mm}
\begin{tabular}{lcccc}
\toprule
Method & Extra params (M) & Peak mem.\ (GB) & Rel.\ mem. & Rel.\ throughput \\
\midrule
From scratch            & --   & 62.0  & 1.00$\times$ & 1.00$\times$ \\
AE           & 2.8  & 69.0  & 1.11$\times$ & 0.91$\times$ \\
DAE          & 2.8  & 69.1  & 1.12$\times$ & 0.91$\times$ \\
DINO                  & 7.9\textsuperscript{$\dagger$} & 128.5 & 2.07$\times$ & 0.41$\times$ \\
SimCLR                & 1.3  & 123.7 & 2.00$\times$ & 0.51$\times$ \\
DiffPT (ours)             & 3.0  & 76.5  & 1.24$\times$ & 0.89$\times$ \\
\bottomrule
\multicolumn{5}{l}{\footnotesize $\dagger$ Trainable student head; DINO also keeps an EMA teacher (+29.2M).}
\end{tabular}
\end{table*}

Across all our experiments, DiffPT shows strong performance, on par or better than SimCLR, which has meticulously fine-tuned augmentations for each dataset, while DiffPT follows common parametrization in the literature. Furthermore, the datasets that SimCLR has the best performance on, i.e., CIFAR-10, Tiny ImageNet, and HAM10000, contain images with the objects being classified typically covering most of the image. For such, arguably simple, classification tasks, the multi-cropping performed by SimCLR has a high likelihood of having the classification object in every crop, thus the contrastive objective directly optimizes for classification. On the other hand, in EyePACS, the biomarkers of diabetic retinopathy are highly localized, possibly just a few pixels (e.g., microaneurysms and small bleeds), and thus it is highly likely that the crops of SimCLR contain patches that indicate different classes (see Figure~\ref{fig:datasets}). DiffPT does not place such location -based priors on the pretraining process, and as such has consistent performance across the range of modalities and tasks.

\subsection{Computational complexity}

Table~\ref{tab:compute} reports the per-step memory cost of each pretraining objective with its own augmentation pipeline. Reconstruction-based objectives are cheap: the auto-encoder, denoising auto-encoder and DiffPT need only 11–24\% more memory than training a classifier from scratch, while being only approximately 10\% slower. This computational advantage stems from the use of asymmetric encoder-decoder architecture, where the decoder is kept simple.

As for the contrastive and self-distillation methods, they require substantially more resources. Both SimCLR and DINO require around 100\% more memory, while the throughput is 50-60\% lower than the baseline classifier. The demands stem from the multi-view strategy utilised by both of these approaches: SimCLR processes two views per image (crops resized to original resolution), while DINO utilises two global ($224\times224$) and four local ($96\times96$) crops. Additionally, DINO requires a forward pass with the teacher network further increasing its latency, but not increasing the memory usage as much, as the teacher is not trained with backpropagation. 

%% file: 06_conclusion.tex
We have benchmarked five self-supervised learning approaches for pretraining classifiers with limited datasets. DiffPT turns out to have strong performance across various experiments ranging from time-series to medical image classification. However, SimCLR outperforms DiffPT for image classification where the discriminative signals cover most of the image area. Interestingly, pretraining with modern self-supervised approaches outperform auto-encoder approaches, with the exception of denoising auto-encoder ranking the second on limited diabetic retinopathy data. Our results suggest that if practitioners have limited classification datasets, and especially if the classification features are poorly understood, DiffPT can be used as an effective classifier pretraining method without extensive hyperparameter tuning. As diffusion models can be easily adapted to different modalities, an interesting direction for future work would be to apply DiffPT to pretrain classification models for multi-modal data.

%% file: 07_appendix.tex
\onecolumn
\section*{Supplementary Materials}
\section{Data augmentations} \label{app:augmentations}

We use two types of data augmentation. The first is a dataset-specific base augmentation, used identically to train the from-scratch classifier baseline, during auto-encoder (AE), denoising auto-encoder (DAE) and DiffPT, and in the fine-tuning stage of every method, SimCLR and DINO included. The second is a view-generating augmentation, used only while pretraining SimCLR and DINO, because their objectives need several stochastically transformed views of each sample. All methods therefore see the same input distribution during fine-tuning, and augmentation differs between methods only where the pretraining objective itself requires it.

\paragraph{Base augmentation.}
Table~\ref{tab:base-aug} lists the base augmentation for each dataset. All
inputs are normalised per channel using training data statistics. For the natural-image datasets we apply the standard random-crop and horizontal-flip. For EyePACS and HAM10000 we additionally use flips about
both axes, arbitrary rotations and mild rescaling. For the time-series
datasets, augmentation consists only of sampling a random fixed-length window
from each record, re-sampled every time the record is drawn. 

\begin{table}[h]
\centering
\small
\caption{Base augmentation.}
\label{tab:base-aug}
\begin{tabularx}{\linewidth}{@{}l l X X@{}}
\toprule
Dataset & Input & Training augmentation & Evaluation \\
\midrule
CIFAR-10      & $32\times32$ & Random crop (4\,px padding); horizontal flip ($p{=}0.5$) & -- \\
Tiny ImageNet & $64\times64$ & Random crop (4\,px padding); horizontal flip ($p{=}0.5$) & -- \\
EyePACS   & $256\times256$ & Random flip (none / horizontal / vertical / both, uniformly); rotation $\in[-180^\circ,180^\circ]$; symmetric padding of up to 5\% per side then random crop back to $256\times256$; scaling $\in[0.9,1.1]$ & -- \\
HAM10000      & $256\times256$ & Resize from $600\times450$, then as EyePACS & Resize \\
\midrule
PTB-XL        & $12\times250$ (2.5\,s @ 100\,Hz) & Random window from the 10\,s record & Sliding windows, 50\% overlap; max over windows \\
WISDM         & $12\times200$ (10\,s @ 20\,Hz) & Random window within the record's valid length & Sliding windows, 50\% overlap; mean over windows \\
\bottomrule
\end{tabularx}
\end{table}

\paragraph{View-generating augmentation for SimCLR and DINO.}
On CIFAR-10 and Tiny ImageNet, for SimCLR each view is produced by RandomResizedCrop (bilinear), a horizontal flip ($p{=}0.5$), colour
jitter (brightness, contrast, saturation, hue) $=(0.8s,0.8s,0.8s,0.2s)$ applied with $p{=}0.8$, and random grayscale with probability $p_g$. For DINO we follow the multi-crop recipe of \citet{caron2021emerging}, producing two global crops and $N_\ell$ local crops (bicubic RandomResizedCrop), each with a horizontal flip ($p{=}0.5$), colour jitter $=(0.4s,0.4s,0.2s,0.1s)$ applied with $p{=}0.8$, and random grayscale with probability $p_g$. Gaussian blur (radius $\in[0.1,2]$) is applied with probability 1.0 to the first global crop, 0.1 to the second and 0.5 to each local crop, and solarisation with probability 0.2 to the second global crop only. Table~\ref{tab:ssl-aug-img} gives the per-dataset settings. 

For EyePACS and HAM10000 we adapt view generation to preserve diagnostically relevant signal. Colour and hue carry clinical information in fundus photographs and in dermoscopic images, so we reduce the colour-jitter strength ($s=0.3$ and $s=0.5$ respectively) and disable random grayscale. Aggressive cropping can remove small lesions entirely, so we also raise the minimum crop scale. These settings are fixed across all training-set sizes.

\begin{table*}[h]
\centering
\small
\caption{SimCLR and DINO view-generation settings for the image datasets.
Crop scales are fractions of the image area passed to RandomResizedCrop;
$s$ is the colour-jitter strength and $p_g$ the grayscale probability.
(see text).}
\label{tab:ssl-aug-img}
\begin{tabular}{@{}l ccc cccc c@{}}
\toprule
& \multicolumn{3}{c}{SimCLR} & \multicolumn{5}{c}{DINO} \\
\cmidrule(lr){2-4} \cmidrule(l){5-9}
Dataset & View size & Crop scale & $s$ / $p_g$
        & Global size & Global scale & Local size & Local scale & $N_\ell$, $s$ / $p_g$ \\
\midrule
CIFAR-10      & 32  & $[0.2,1]$ & 1.0 / 0.2 & 32  & $[0.5,1]$  & 24 & $[0.2,0.6]$ & 6, 1.0 / 0.2 \\
Tiny ImageNet & 64  & $[0.2,1]$ & 1.0 / 0.2 & 64  & $[0.35,1]$ & 32 & $[0.1,0.4]$ & 6, 1.0 / 0.2 \\
EyePACS   & 256 & $[0.5,1]$ & 0.3 / 0   & 224 & $[0.6,1]$  & 96 & $[0.3,0.6]$ & 4, 0.3 / 0 \\
HAM10000$^\dagger$ & 256 & $[0.5,1]$ & 0.5 / 0 & 224 & $[0.6,1]$ & 96 & $[0.3,0.6]$ & 4, 0.5 / 0 \\
\bottomrule
\end{tabular}

\end{table*}

For time-series data, image augmentations such as colour jitter, grayscale, blur and solarisation have no meaningful counterpart for multichannel physiological or inertial signals. For PTB-XL and WISDM we therefore replace them with a time-series augmentation pipeline. Each view is produced independently from the same record by (i)~a random temporal crop of fixed length $L$ (the SimCLR view length, or the DINO global or local crop length; (ii)~additive Gaussian noise $\tilde{x}=x+\epsilon$, $\epsilon\sim\mathcal{N}(0,\sigma^2 I)$, on the z-normalised signal; and (iii)~masking one contiguous segment of width $w\sim\mathcal{U}\{0,\dots,\lfloor 0.1L\rfloor\}$ to zero (the per-channel training mean) across all channels simultaneously. Crops are never resized. The network's global pooling handles DINO's shorter local crops. For PTB-XL we set $\sigma=0.15$. For WISDM we use a milder $\sigma=0.1$ relative to the unit-variance channels. Table~\ref{tab:ssl-aug-ts} summarises the settings.

\begin{table}[h]
\centering
\small
\caption{SimCLR and DINO view generation for the time-series datasets. Window
lengths are in samples, with durations in parentheses.}
\label{tab:ssl-aug-ts}
\begin{tabular}{@{}l cc@{}}
\toprule
& PTB-XL & WISDM \\
\midrule
Sampling rate                     & 100\,Hz & 20\,Hz \\
SimCLR view length                & 250 (2.5\,s) & 200 (10\,s) \\
DINO global crops                 & $2\times250$ (2.5\,s) & $2\times200$ (10\,s) \\
DINO local crops                  & $6\times125$ (1.25\,s) & $6\times100$ (5\,s) \\
Gaussian noise $\sigma$           & 0.15 & 0.10 \\
Time-mask width (max), global / local & 25 / 12 & 20 / 10 \\
\bottomrule
\end{tabular}
\end{table}